# Multimodal Sentiment Analysis based on Multi-channel and Symmetric Mutual Promotion Feature Fusion

Wangyuan Zhu, Jun Yu*, *Member, IEEE*

*Abstract*—**Multimodal sentiment analysis is a key technology in the fields of human-computer interaction and affective computing. Accurately recognizing human emotional states is crucial for facilitating smooth communication between humans and machines. Despite some progress in multimodal sentiment analysis research, numerous challenges remain. The first challenge is the limited and insufficiently rich features extracted from single modality data. Secondly, most studies focus only on the consistency of inter-modal feature information, neglecting the differences between features, resulting in inadequate feature information fusion. In this paper, we first extract multi-channel features to obtain more comprehensive feature information. We employ dual-channel features in both the visual and auditory modalities to enhance intra-modal feature representation. Secondly, we propose a symmetric mutual promotion (SMP) inter-modal feature fusion method. This method combines symmetric cross-modal attention mechanisms and self-attention mechanisms, where the cross-modal attention mechanism captures useful information from other modalities, and the self-attention mechanism models contextual information. This approach promotes the exchange of useful information between modalities, thereby strengthening inter-modal interactions. Furthermore, we integrate intra-modal features and inter-modal fused features, fully leveraging the complementarity of inter-modal feature information while considering feature information differences. Experiments conducted on two benchmark datasets demonstrate the effectiveness and superiority of our proposed method.**

*Index Terms*—**Sentiment Recognition, Multimodal, Feature Extraction, Feature Fusion, Attention Mechanism.**

## I. INTRODUCTION

IN the field of human-computer interaction, accurately recognizing human sentiment is crucial for facilitating smooth communication between humans and machines, and effectively adapting to users' emotional fluctuations is key to enhancing user experience. Compared to single-modal sentiment recognition systems, multimodal sentiment recognition systems integrate data from different channels, enabling a more comprehensive capture and interpretation of sentiment information. While some progress has been made in multimodal sentiment recognition research, numerous challenges remain. Firstly, previous studies often focused on single-modal data such as visual, auditory, or textual, resulting in a limited and insufficiently diverse set of features extracted within each modality, thereby constraining the accuracy and stability of the model. Secondly, although some research attempts to improve recognition accuracy through multimodal approaches, most only consider the consistency of feature information between modalities, neglecting the differences between features and leading to insufficient feature fusion. When performing multimodal feature fusion, there are significant differences in emotional features between different modalities, and these features may have complex interactions. For instance, a person might use sarcastic language while displaying a relaxed facial expression, yet their voice might carry a noticeable tone of anger. These two contradictory emotional polarity features, if simply concatenated, could lead to information confusion and redundancy, thereby affecting the final model's recognition performance. However, despite the conflicting nature of emotional expressions across different modalities, there is still a certain degree of correlation among them. For example, one aspect of emotional expression might influence another, such as the tone of voice reflecting in the speaker's facial expression, thus providing additional emotional information. Therefore, in multimodal feature fusion, the challenge lies in how to capture the correlations between different modalities while preserving the intrinsic semantic information of each modality.

To address the issue of limited features within single modalities affecting model accuracy and stability, this thesis proposes extracting multi-channel features to enrich intra-modal features, thereby improving the accuracy of single-modal sentiment recognition models. For the visual modality, global spatial feature extraction based on the ResNet18 network and local expression feature extraction based on AUs are proposed, integrating these features to construct a comprehensive visual feature. For the auditory modality, traditional handcrafted feature extraction based on MFCC and deep learning feature extraction based on Wav2Vec2.0 are proposed, combined to obtain higher-quality auditory features. Additionally, an attention mechanism is introduced into the bidirectional long short-term memory network, which captures temporal correlations of feature sequences while enhancing the model's focus on key information.

To address the issue of insufficient feature fusion due to significant differences in feature information between different modalities, this thesis proposes a feature fusion method based on Symmetric Mutual Promotion. This method combines symmetric cross-modal attention mechanisms and self-attention mechanisms, where the cross-modal attention mechanism captures useful information from other modalities and the self-attention mechanism models contextual information. This

Wangyuan Zhu and Jun Yu are with the Department of Automation, University of Science and Technology of China, Hefei, Anhui 230022, China. E-mail: zhuwangyuan@mail.ustc.edu.cn; harryjun@ustc.edu.cn.
* is the corresponding author.

method promotes the exchange of useful information between modalities, strengthening their interaction. Furthermore, features within each modality and fusion features between modalities are integrated, fully utilizing the complementary nature of feature information between modalities while considering differences in feature information.

The contributions can be summarized as follows:

- We propose a method to enrich intra-modal feature information by extracting multi-channel features, combining traditional handcrafted features with deep learning features to obtain more comprehensive feature information.
- We propose a symmetric mutual promotion-based inter-modal feature fusion method. This approach combines a symmetric cross-modal attention mechanism with a self-attention mechanism, where the cross-modal attention mechanism captures useful information from other modalities, while the self-attention mechanism models contextual information.
- We conducted extensive experiments on two publicly available datasets, CMU-MOSI and CH-SIMS. The experimental results demonstrate that our proposed method outperforms the baselines.

## II. RELATED WORK

### A. Single-model Sentiment Analysis

For visual inputs, facial expression recognition has been a hot topic in sentiment analysis research, attracting widespread attention in academia. As early as 1978, Ekman et al. [1] pioneered the Facial Action Coding System (FACS), summarizing six basic human emotions: joy, sadness, anger, fear, surprise, and disgust. Depending on the continuity of input data, facial expression recognition can be further categorized into static facial images and dynamic video sequences. Static facial images refer to face images without temporal information. In early research, traditional handcrafted feature extraction methods such as Local Binary Patterns (LBP) [2], Scale-invariant Feature Transform (SIFT) [3], Histogram of Oriented Gradients (HOG) [4], and Gabor wavelet coefficients [5] were widely used for facial expression analysis and recognition. Xia et al. [6] proposed weighted fusion features to represent facial expressions. They first used an operator called Multi-Scale Block Local Binary Patterns Uniform Histogram (MB-LBPUH) to filter facial images and then concatenated MB-LBPUH with HOG to form a new feature representation for facial expressions. With the rise of neural networks, various from-scratch or fine-tuned pretrained CNN models have been applied to facial expression recognition tasks, such as AlexNet [7], VGGNet [8], GoogleNet [9], ResNet [10], etc. Sun et al. [11] analyzed the optical flow changes between the peak moments of expressions and neutral expressions to capture the temporal dimension of facial expressions, complemented by grayscale images of emotional expressions to provide spatial information.

For speech input, traditional methods often rely on manually crafted Low-Level Descriptors (LLD) to extract emotional features from speech signals. These features include rhythmic features, voice quality features, and spectral features, among others [12], [13]. Rhythmic features involve aspects such as pitch, energy, amplitude, and duration [14], while voice quality features may include resonance peaks, spectral energy distribution, and glottal features [15]. For instance, Schmitt et al. [16] used the Mel-Frequency Cepstral Coefficients (MFCC)-based Bag of Audio Words (BoAW) approach as a feature representation and employed Support Vector Machine (SVM)-based regression to predict time-continuous arousal and valence emotions. Cheng et al. [17] proposed a Two-Layer Fuzzy Multiple Random Forests (TLFMRF) model, which first utilized OpenSMILE to extract the ComParE-2013 feature set from speech signals. Then, they applied the fuzzy C-means clustering method to classify high-dimensional speech features and finally used multiple random forests to identify different emotion categories. With the advancement of deep learning technologies, a series of deep learning-based algorithms have been applied to SER tasks to extract more complex and abstract speech features. Zhang et al. [18] utilized multidimensional CNNs, including 1D, 2D, and 3D CNNs, to extract deep multimodal segment-level feature representations. They adopted average pooling and decision-level fusion strategies to integrate discourse-level classification results, demonstrating that deep learning models can learn complementary multimodal speech representations effectively.

For text input, early methods relied on Bag-of-Words (BoW) models [19] to extract handcrafted emotional features from text. This model learns vector representations of text by computing the frequency of each word in the document. Colneric et al. [20] studied two methods to convert raw text into BoW models for sentiment analysis on Twitter. One method was a standard BoW model, and the other was a normalized BoW model aimed at reducing model complexity by lowering the dimensionality of features. To overcome the limitations of BoW models, Blei et al. [21] proposed Latent Dirichlet Allocation (LDA), a probabilistic process-based method to extract latent topics from documents, thereby revealing the semantic structure behind the text. With the advancement of deep learning technologies, text feature extraction has shifted towards word embedding techniques. These techniques learn continuous vector representations of words to capture general semantic and contextual information. Word2vec [22] and GloVe [23] are two typical word embedding methods that learn vector representations of words through different training strategies, allowing them to capture fine-grained syntactic and semantic regularities between words.

### B. Multi-modal Sentiment Analysis

In order to more accurately simulate human comprehensive perception during sentiment analysis and extract and integrate various forms of emotional expression information, effective integration of information from different expressive channels has become crucial for improving recognition accuracy. In the field of multimodal sentiment analysis, a major challenge is how to effectively parse and integrate information from different modalities to derive precise sentiment analysis results. When performing multimodal fusion, it is necessary to consider the variability within each modality and the correlations between different modalities, ensuring maximal utilization of complementarity and synergistic effects among various modal data.

Early research primarily focused on the fusion of two modalities. For example, Zhang et al. [24] proposed a hybrid deep learning (DBN) method for visual and speech sentiment analysis, where they first used CNN and 3D-CNN to extract visual and speech features respectively, and then merged these features using a DBN model followed by emotion classification using a linear SVM classifier. Krishna et al. [25] developed a novel approach based on cross-modal attention and 1D CNNs, where they extracted advanced features from raw speech signals using an audio encoder and semantic information from text using a text encoder, then interactively fused audio and text sequences using cross-modal attention networks, and finally performed emotion classification using fully connected layers and softmax. Kumar et al. [26] considered social data in sentiment analysis by integrating text and visual modalities, and introduced a deep learning classification model named Context ConvNet-SVM$_{BoVW}$, which includes discretization, text and image analysis, and decision modules, enhancing sentiment analysis capabilities across different modalities in online social media.

In recent years, multimodal sentiment analysis research has gained increasing attention. Dai et al. [27] proposed an end-to-end multimodal sparse model (MESM), which utilized pretrained VGG16 models for audio and visual feature extraction, modeled temporal information using Transformer, and introduced cross-modal sparse CNN blocks to reduce computational costs, ultimately achieving classification results through feedforward networks (FFN). Mittal et al. [28] proposed an innovative M3ER multimodal sentiment analysis model, particularly designing a multiplication fusion layer. This model aimed to identify and learn more reliable information from each modality, effectively suppressing less relevant modality information considering specific sample conditions. They used popular LLD features (such as MFCC, pitch, and glottal source parameters) for speech features, facial action units and landmarks for visual features, and pretrained GloVe models for text features. They then executed a multiplication modality fusion approach, followed by emotion classification through FC and softmax layers. However, their approach fundamentally relied on early feature-level fusion techniques, which, while simplifying the model to some extent, often sacrificed robustness.

Zhao et al. [29] proposed a prompt learning-based multimodal sentiment analysis pretrained model (Memobert). Memobert was capable of learning joint multimodal feature representations from large-scale unlabeled video data via self-supervised learning. The model consisted of three specific modality encoder blocks, three specific modality embedding blocks, and a multilayer cross-modal Transformer block. For audio modality, frame-level acoustic features were obtained using pretrained Wav2Vec2.0. For visual modality, facial expression-related features were extracted using a pretrained dense network. For text modality, textual features were obtained using pretrained BERT.

Despite the progress made in multimodal sentiment anal-ysis methods, most researchers have focused only on the consistency of intermodal feature information, neglecting the differences between features, thereby insufficiently integrating feature information.

## III. The Proposed Method

The overall structure is shown in the Fig. 1. It is mainly divided into an intra-modal feature extraction module and an inter-modal feature fusion module, where the prediction module further integrates branch-specific feature information and modal fusion information to obtain more accurate prediction results.

### A. Multi-channel feature extraction

*1) Videos features:* In the preprocessing stage, we have extracted continuous face images from the video, allowing us to directly extract facial expression features from these images.

**ResNet18**: The Convolutional Neural Network (CNN) ResNet18, introduced by [30], is renowned for its exceptional ability to extract features from images. In order to enhance its performance on facial datasets, a ResNet18 network pretrained on AffectNet [31] is utilized to extract global spatial features from facial images. The features before the final fully connected layer are averaged to obtain a 512-dimensional feature vector. This improvement further enhances the network's capability to accurately extract relevant facial features.

**AUs**: The Facial Action Coding System (FACS) [1] is a comprehensive method for objectively coding facial expressions. In FACS, Action Units (AUs) correspond to specific facial muscles. Each AU has two dimensions: the first dimension indicates detection, with 0 indicating absence and 1 indicating presence. The second dimension represents intensity, ranging from 0 to 1. We utilize OpenFace2.0 [32] to extract the detection and intensity of 17 AUs relevant to facial expressions. Ultimately, each facial image receives a 34-dimensional feature embedding.

*2) Audios features:* Before extracting the audio features, we normalized all audio files to -3 dB and converted them to a format of 16 kHz, 16-bit mono.

**Wav2Vec2.0**: Self-supervised pre-trained Transformer models have garnered considerable attention in the field of computer audition. A prominent example of such a foundational model is Wav2Vec2.0 [33], which is frequently employed for Speech sentiment analysis (SER) [34]. Given that all subchallenges are emotion-related, we leverage Wav2Vec2.0, specifically a large version fine-tuned on the MSP-Podcast [35] dataset, for continuous sentiment analysis. We derive audio signal features by averaging the representations from the final layer of the model, resulting in a 768-dimensional embedding.

**MFCC**: Mel-Frequency Cepstral Coefficients (MFCC) [36] is a feature extraction method used in speech processing and speech recognition. MFCC captures variations in pitch and timbre of speech signals by simulating human ear perception. The extraction process involves segmenting the speech signal into short frames, applying windowing, performing a Fast Fourier Transform (FFT) to obtain the spectrum, converting the spectrum to the mel scale using a mel filter bank, taking the logarithm of the mel spectrum, and finally applying a Discrete Cosine Transform (DCT) to obtain the MFCC features.

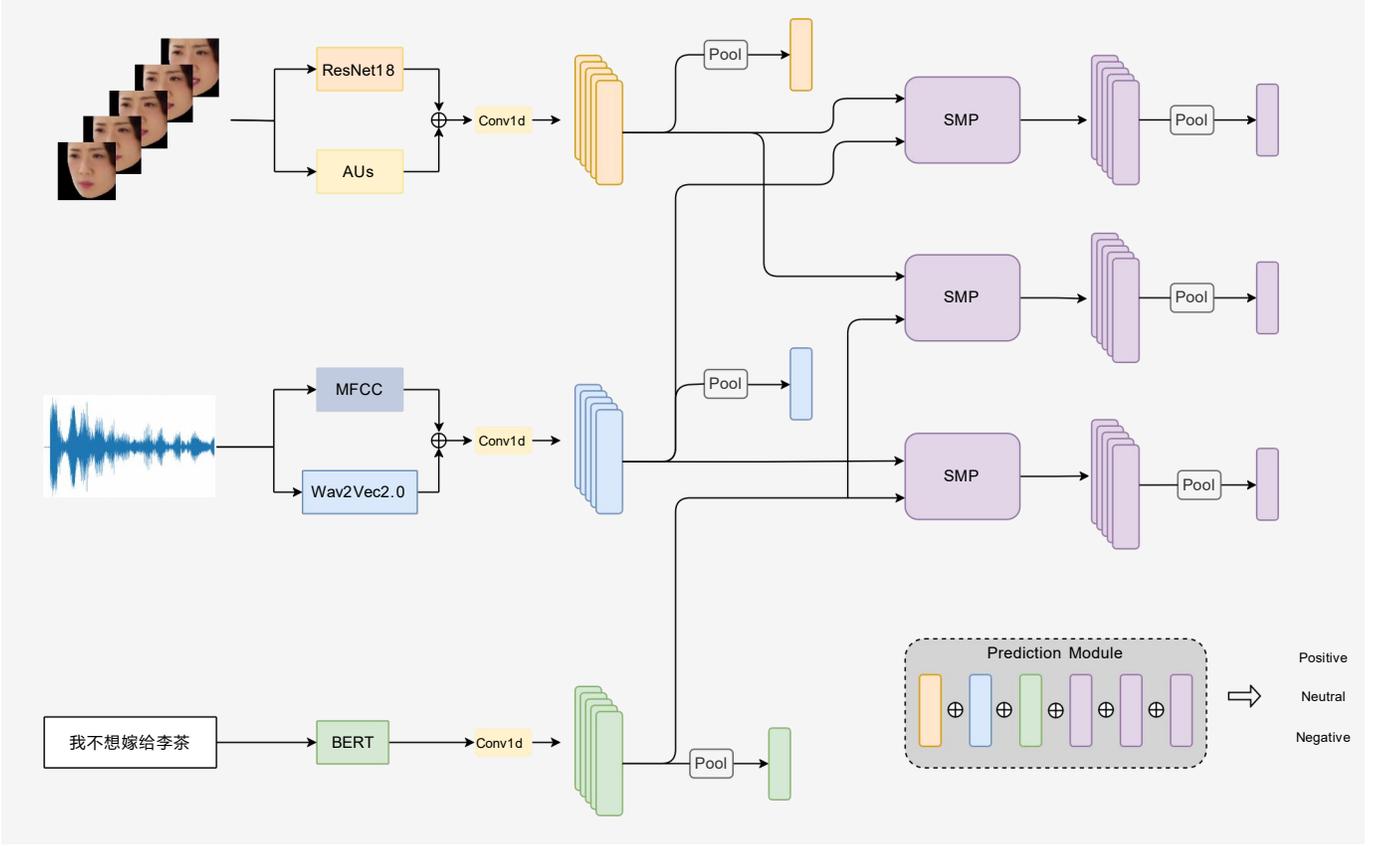

Fig. 1. Structure Diagram of the sentiment analysis Model Based on Vision, Audio, and Text

*3) Text features:* For the original text data, this paper first uses the word embedding technology provided by Tencent to encode the text and obtain a vector representation to avoid the Out-of-Vocabulary (OOV) problem.

**BERT**: Bidirectional Encoder Representations from Transformers (BERT) [37] is a pre-trained language model developed by Google. BERT is designed to understand the context of a word in search queries and text by considering both the words that come before and after it. Unlike previous models that read text sequentially, BERT uses a bidirectional approach to process text, allowing it to capture richer, more nuanced information. This capability makes BERT highly effective for various natural language processing tasks such as question answering, sentiment analysis, and language translation.

### B. Symmetry mutually promotes feature fusion module

Considering the symmetry between the visual-audio, visual-text, and audio-text modalities, the fusion methods for these three modality pairs are structurally similar. Therefore, we first introduce the Symmetric Mutual Promotion (SMP) feature fusion module proposed in this paper. The structure of this module is shown in Fig. 2. It primarily explores the intrinsic correlations between the elements of the two input feature sequences through two symmetric cross-modal attention mechanisms. This approach allows for mutual data flow, promoting the exchange of useful information through the attention mechanism, thereby achieving the effect of mutual promotion.

Now, let's analyze the structure of the SMP model from the visual-audio perspective. The model inputs are visual features $X_v \in R^{T_v \times d_v}$ and audio features $X_a \in R^{T_a \times d_a}$. During the preprocessing stage, data of the same time length is uniformly extracted from both video and audio, so $T_v$ and $T_a$ are equal. In subsequent calculations, we use $T$ to represent the time steps of the feature sequences. Additionally, for convenient feature fusion in the next step, a linear layer is concatenated to the input features, mapping them to the same spatial dimension $d$. First, positional encoding is applied to the input feature sequences to differentiate the elements at various positions and capture the sequential relationships within the series. The calculation method is shown in equation 2, where $PE \in R^{T \times d}$ represents the sine-cosine positional encoding.

$$X_v = X_v + PE_v \quad (1)$$

$$X_a = X_a + PE_a \quad (2)$$

The SMP module consists of two symmetrical structures, referred to here as the left branch $LSMP(X_v, X_a)$ and the right branch $RSMP(X_a, X_v)$, indicating the direction of feature information flow. Taking $LSMP(X_v, X_a)$ as an example, its calculation formula is as follows:

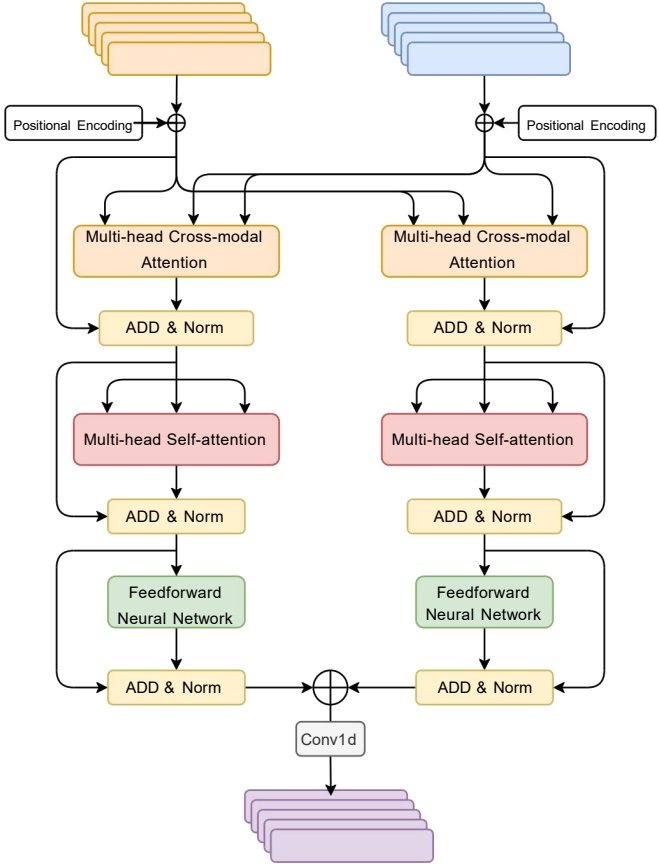

Fig. 2. Structure of the SMP Module

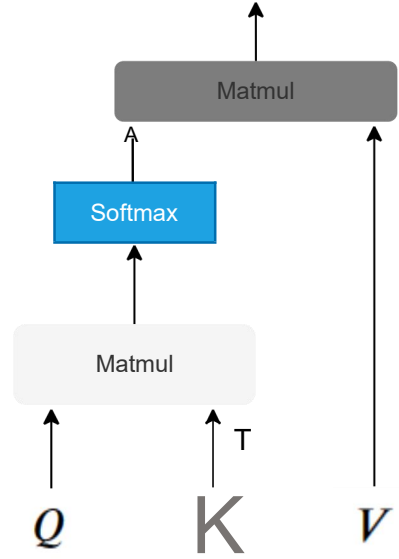

Fig. 3. Diagram of Attention Mechanism

$$\text{LSMP}'(X_v, X_a) = \text{LN}(\text{MHCA}(X_v, X_a) + X_v) \quad (3)$$

$$\text{LSMP}''(X_v, X_a) = \text{LN}(\text{MHSA}(\text{LSMP}'(X_v, X_a)) \quad (4)$$

$$+ \text{LSMP}'(X_v, X_a)) \quad (5)$$

$$\text{LSMP}(X_v, X_a) = \text{LN}(\text{FFN}(\text{LSMP}''(X_v, X_a)) \quad (6)$$

$$+ \text{LSMP}''(X_v, X_a)) \quad (7)$$

In the formula, MHSA and MHCA represent Multi-Head Self-Attention and Multi-Head Cross-Modal Attention mechanisms, respectively. Both are attention mechanisms based on the Transformer [38] structural components. The structure of this mechanism is shown in Fig. 3.

Self-Attention (SA) is a key component in the Transformer model. It utilizes a scaled dot-product attention mechanism to model global dependencies within sequences, enabling efficient parallel computation and capturing long-range dependencies. In SA, given an input feature sequence $H_n \in \mathbb{R}^{T_n \times d}$ with $T_n$ tokens, each token represented by a $d$-dimensional vector, multiple matrices are derived through linear transformations: query matrix $Q_n = H_n W_Q$, key matrix $K_n = H_n W_K$, and value matrix $V_n = H_n W_V$, where $W_Q$, $W_K$, and $W_V \in \mathbb{R}^{d \times d}$. The computation formula for $\text{SA}(H_n)$ is as follows:

$$\text{SA}(H_n) = \text{softmax}(\frac{Q_n K_n^T}{\sqrt{d_k}}) V_n \quad (8)$$

In this context, $\sqrt{d_k}$ serves as a scaling factor applied to the dot product result to stabilize its variance. The softmax function computes attention weights, normalizing the dot product $Q_n K_n^T$ into an attention distribution. Finally, these attention weights are multiplied by the value matrix $V_n$, resulting in a weighted output. Essentially, this process allocates weights to positions in the value matrix $V_n$ based on the similarity between the query matrix $Q_n$ and the key matrix $K_n$. Thus, the model gains the ability to capture and model the interdependencies among different positions within the input sequence.

Typically, Cross-modal Attention (CA) takes as input two modal feature sequences, where the query matrix $Q$ comes from the target modality $t$, and the key and value matrices $K$ and $V$ originate from the source modality $s$. They can be represented as $Q_t = H_t W_Q$, $K_s = H_s W_K$, and $V_s = H_s W_V$. Through this mechanism, CA provides a potential representation from the source modality $s$ to the target modality $t$, with the computation formula as follows:

$$\text{CA}(H_t, H_s) = \text{softmax}\left(\frac{Q_t K_s^T}{\sqrt{d_k}}\right) V_s \quad (9)$$

Specifically, the $\text{MHCA}(X_v, X_a)$ module receives the query matrix $Q$ from the visual modality, and the key and value matrices $K$ and $V$ from the audio modality. Through cross-modal attention mechanism, it captures a potential representation from the audio modality to the visual modality and extracts useful information. Similarly, $\text{MHCA}(X_a, X_v)$ receives the key and value matrices $K$ and $V$ from the visual modality and the query matrix $Q$ from the audio modality. Through cross-modal attention mechanism, it captures a potential representation from the visual modality to the audio modality and obtains useful information. Subsequently, the outputs of each modality undergo self-attention mechanism to effectively capture long-range dependencies between different positions in the feature sequences, and dynamically compute attention weights for each position to model global semantic correlations.

Here, LN denotes Layer Normalization. In Layer Normalization, each sample's features are normalized across all features of that sample. Given an input $x$, the normalized output $y$ is computed as:

$$y = \frac{x - \mu}{\sqrt{\sigma^2 + \epsilon}} \qquad (10)$$

where $\mu$ is the mean of the input sample features, $\sigma^2$ is the variance, and $\epsilon$ is a small number to prevent division by zero. By standardizing the input data, the mean is centered around zero and the variance is close to one, thereby reducing gradient vanishing or exploding issues, accelerating convergence, and improving training stability and generalization capability of the model.

In the equations, FFN refers to Feedforward Neural Network, consisting of two fully connected layers with a non-linear activation function. The process first performs linear mapping through the first layer, followed by a non-linear activation such as ReLU function, and then completes the final linear transformation through the second layer to obtain the output result. The computation process of FFN can be represented as:

$$FFN(x) = ReLU(xW_1 + b_1)W_2 + b_2 \qquad (11)$$

where $x$ represents the feature vector obtained after the attention mechanism. $W_1$ and $b_1$ correspond to the weight matrix and bias term of the first fully connected layer, respectively. Similarly, $W_2$ and $b_2$ represent the weight and bias of the second fully connected layer. The activation function ReLU refers to Rectified Linear Unit function defined as $ReLU(x) = \max(0, x)$. By applying non-linear transformations and remapping features on the output of the attention layer, the model's expressive power and depth are significantly enhanced.

Similarly, the computation process of the right branch $RSMP(X_a, X_v)$ is as follows:

$$RSMP^{'}(X_a, X_v) = LN(MHCA(X_a, X_v) + X_a) \qquad (12)$$

$$RSMP^{''}(X_a, X_v) = LN(MHSA(RSMP^{'}(X_a, X_v)) \qquad (13)$$

$$+ RSMP^{'}(X_a, X_v)) \qquad (14)$$

$$RSMP(X_a, X_v) = LN(FFN(RSMP^{''}(X_a, X_v)) \qquad (15)$$

$$+ RSMP^{''}(X_a, X_v)) \qquad (16)$$

Finally, by concatenating the outputs of the left and right branches in the spatial dimension and applying a 1D convolution, the SMP outputs fused features $F_{va} \in R^{T \times d}$ are obtained, with the formula as follows:

$$F_{va} = SMP(X_v, X_a) \qquad (17)$$

$$= Conv1d(Concatenate(LSMP(X_v, X_a) \qquad (18)$$

$$+ RSMP(X_a, X_v))) \qquad (19)$$

This symmetric mutual promotion attention mechanism enables both sides to effectively capture and utilize each other's relevant information, thereby facilitating effective information exchange and interaction between different modalities. By integrating information from both sides, this mechanism further refines richer interactive features, enhancing the model's understanding and processing capabilities of multi-modal data.

Compared to bimodal sentiment analysis models, multimodal sentiment analysis models maintain consistency in overall architecture, including three branches for visual, audio, and text modalities, along with three symmetric mutual promotion (SMP) feature fusion modules and a prediction module. Each SMP module takes inputs corresponding to pairs of visual-audio, visual-text, and audio-text modalities, aiming to deeply explore and integrate interactions and influences between different modalities. After feature fusion, the model extracts global representations of each modality fusion feature through pooling operations. The specific calculation methods are as follows:

$$F_{va} = Pool(SMP(X_v, X_a)) \qquad (20)$$

$$F_{vt} = Pool(SMP(X_v, X_t)) \qquad (21)$$

$$F_{at} = Pool(SMP(X_a, X_t)) \qquad (22)$$

For the prediction module, it integrates outputs from the visual, audio, and text branches, along with the outputs of the three modality fusion features. It uses linear mapping through fully connected layers to obtain the final sentiment analysis prediction:

$$y = softmax(FC(Concatenate(F_v, F_a, F_t, F_{va}, F_{vt}, F_{at}))) \qquad (23)$$

Here, $y$ represents the prediction result with a dimension of 3 (for 3-classification tasks), indicating probabilities for positive, neutral, and negative emotions.

## IV. EXPERIMENTS

### A. Datasets

In this section, we evaluate and analyze the performance of the model using CMU-MOSI [39] and CH-SIMS [40] datasets. Both datasets provide three modalities of data for sentiment analysis: Text (T), Audio (A), and Visual (V).

TABLE I
DATASET SETTINGS

| Dataset | Training set | Valid set | Testing set | Summary |
|---------|-------------|-----------|-------------|---------|
| CMU-MOSE | 1284 | 229 | 686 | 2199 |
| CH-SIMS | 1368 | 456 | 457 | 2281 |

**CMU-MOSI:** The CMU-MOSI dataset comprises $2,199$ video clips, primarily segmented from the reviews of 89 speakers on 93 movies. Each video clip has been manually annotated with an emotional score ranging from -3 to 3, indicating the speaker's sentiment. A score greater than 0 signifies positive sentiment, while a score less than 0 indicates negative sentiment. The greater the absolute value of the emotional score, the stronger the sentiment expressed, for details see Table I.

**CH-SIMS:** The CH-SIMS dataset is a Chinese multimodal emotion analysis dataset that includes unified multimodal annotations as well as independent unimodal annotations. This dataset was primarily collected from 60 Chinese movies, TV shows, and variety shows. The criterion for segmenting original video samples was based on each video segment having a length not less than 1 second and not more than 10 seconds. In total, 2281 short video clips were obtained through this process, for details see Table I. Additionally, it was ensured that each video segment contains only one speaker, excluding any other characters besides the current speaker in the video, These segments all exhibit emotional expressions and are exclusively in Mandarin Chinese, reflecting everyday language use. Among these 2281 data samples, the average duration of video and audio is 3.67 seconds, while the average length of text is 15 words per sentence.

**Preprocessing:** Videos, as a form of multimedia data, contain richer information and are more closely aligned with real-life scenarios. They not only capture temporal changes in facial expressions but also facilitate better model application in real-world detection and recognition tasks. However, several challenges arise from factors such as occlusions, strong lighting conditions, similar backgrounds, and significant facial movements like head turns, which can make it difficult to accurately detect faces in multiple consecutive frames extracted from videos. These factors increase the complexity of extracting facial expression features from videos. Additionally, the continuity of video data may involve movement of individuals within the frame, further complicating the task of face alignment. Therefore, the primary goal of preprocessing is to accurately detect faces and align them across consecutive video frames to ensure consistent input data size for the model. For raw video files, the initial step involves using tools like OpenCV to convert video content into a series of continuous image frames. Each frame is then subjected to face detection; frames containing detected faces are retained, while frames without detected faces are discarded. Specifically, this study utilizes the Multi-task Cascaded Convolutional Networks (MTCNN) [41], a multi-task convolutional neural network architecture. MTCNN comprises three cascaded CNN modules: the Proposal Network (P-Net), a compact fully convolutional network designed to efficiently generate face candidate regions; the Refine Network (R-Net), which further refines and filters initial face candidate boxes from P-Net to significantly improve face detection accuracy; and the Output Network (O-Net), another efficient CNN responsible for detecting facial landmarks such as eyes, nose, and mouth on precise face candidate boxes generated by R-Net. O-Net's role is critical for achieving precise face alignment. Following this meticulous detection process, the model uses the coordinates of facial landmarks provided by MTCNN to accurately remove background interference, crop the faces to appropriate sizes, and standardize each image to a uniform size of 112x112 pixels.

In CH-SIMS dataset, the raw video data undergo preprocessing to generate a series of consecutive facial images, as illustrated in Fig. 4. The left side of the arrows in the figure shows the unprocessed raw video, while the right side displays a sequence of clear facial images after cropping and adjustment. These consecutive facial images provide high-quality visual inputs for subsequent sentiment analysis tasks, ensuring accuracy and reliability in recognition.

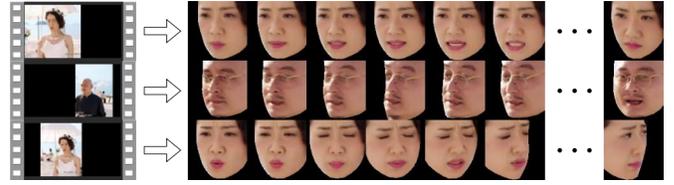

Fig. 4. Preprocessed consecutive facial images, left: original Videos, right: continuous face images

### B. Baselines

The following are detailed descriptions of these baseline models:

**Early Fusion LSTM ( EF-LSTM ) [42]:** This model first concatenates the inputs from three modalities and then uses LSTM to capture long-distance dependencies in the sequence. This model aims to combine the information from multiple modalities early on and then process them to achieve interaction between modalities.

**Later Fusion DNN ( LF-DNN ) [43]:** In contrast to EF-LSTM, this model learns single-modality features first and then concatenates them before classification. The main idea of this model is to process the information from each modality first and then combine them to achieve interaction between modalities.

**Memory Fusion Network (MFN) [44]:** This model considers view-specific and cross-view interactions and models them continuously over time using a special attention mechanism, summarizing them over time through multiview gated memory. MFN requires word-level alignment of three modalities, but since reliable Chinese corpus alignment tools are lacking, CTC is used as an alternative method in this experiment.

**Tensor Fusion Network (TFN) [45]:** This model explicitly models view-specific and cross-view dynamics by creating a multidimensional tensor that spans single-modality, double-modality, and triple-modality interactions. The main idea of this model is to combine information from different modalities and represent it as a multidimensional tensor, and then use tensor decomposition to learn interaction between modalities.

**Multimodal Transformer (Mult) [46]:** This model uses directional pairwise cross-modality attention to achieve interaction between multimodal sequences across different time steps and potentially adapt one modality to the flow of another. The main idea of this model is to use multihead self-attention mechanisms to process information from different modalities and use cross-modality attention mechanisms to achieve interaction between modalities.

### C. Evaluation Metrics

Since emotion intensity prediction is primarily a regression task, commonly used evaluation metrics include Mean Absolute Error (MAE) and Pearson Correlation Coefficient (Corr). Researchers also convert continuous scores into various discrete categories and report classification accuracy. We report seven-level accuracy (Acc-7), five-level accuracy (Acc-5), binary accuracy (Acc-2), and F1-score. For CH-SIMS, we report five-level accuracy (Acc-5), three-level accuracy (Acc-3), and binary accuracy (Acc-2). For all metrics except MAE, higher values indicate better performance.

### D. Experimental Setup

This study implemented all proposed models using the Py-Torch framework, with details of the development environment provided in Table II. Uniform sizing in the time dimension of input sequences was set to 32 for videos, audio, and text. Samples smaller than 32 dimensions were zero-padded, while samples larger than 32 dimensions were uniformly sampled to extract 32 dimensions. During the training phase, the models were trained using a single GPU resource. The Adam optimizer was employed to optimize network parameters. To ensure adequate convergence of the models, training was conducted over 100 epochs, starting with an initial learning rate of $1 \times 10^{-5}$. The batch size was set to 64.

TABLE II
DEVELOPMENT ENVIRONMENT CONFIGURATION LIST

| Experimental Configuration | Parameter description |
|---|---|
| Operating system | Ubuntu 20.04.3 LTS |
| Internal memory | 24 GB |
| CPU | Intel(R) Xeon(R) Gold 6248R @ 3.00GHz |
| GPU | NVIDIA GeForce RTX 3090 |
| CUDA | 11.3 |
| cuDNN | 7.0 |
| Pytorch | 1.12.1 |

### E. Experimental Results

*1) Multimodal results on CMU-MOSI dataset:* We conducted multimodal experiments on the CMU-MOSE dataset using visual, speech, and text inputs. The experimental results are presented in Table III. Our proposed method ("Ours") excelled in several key metrics. Specifically, the "Ours" model achieved the highest scores in binary classification accuracy (Acc-2), F1 score, and Pearson correlation coefficient (Corr), with values of 83.44%, 83.21%, and 72.38%, respectively. This indicates high precision and consistency in the binary classification task. Although the "Ours" model did not surpass the best-performing Mult model in five-class (Acc-5) and seven-class (Acc-7) accuracies, its scores were still close to the best, with 41.23% and 35.42%, respectively. In terms of mean absolute error (MAE), the "Ours" model also performed well, with a score of 88.42, which is very close to the Mult model's best score of 87.99. Overall, our proposed method outperformed the baselines across several evaluation metrics, demonstrating its effectiveness and superiority in multimodal sentiment analysis tasks.

*2) multimodal results on CH-SIMS dataset:* Similarly, we conducted multimodal experiments on the CH-SIMS dataset, with results shown in Table IV. The results indicate that our method outperforms other models on several key metrics. Specifically, our model achieves an accuracy of 80.12% on binary classification (Acc-2), which is several percentage points higher than the second-best model. Additionally, it achieves the highest accuracy of 40.12% on five-class classification (Acc-5), and a correlation coefficient (Corr) of 59.32%, indicating better consistency with actual data. While our model's F1 score (79.38%) is slightly lower than the Mult model's highest score of 79.66%, it remains highly competitive. Furthermore, our model's Mean Absolute Error (MAE) is 44.12, slightly higher than TFN model's best score of 43.22 but still superior to most other models. These results highlight the robustness and effectiveness of our proposed method in multimodal sentiment analysis tasks on the CH-SIMS dataset.

TABLE III
THE MULTIMODAL EXPERIMENTAL RESULTS ON THE CMU-MOSE DATASET ARE COMPARED, WITH "OURS" REPRESENTING THE METHOD USED IN THIS PAPER. WE RAN EACH MODEL FIVE TIMES AND REPORTED THE AVERAGE RESULTS. FOR ALL METRICS EXCEPT MAE, HIGHER VALUES INDICATE BETTER PERFORMANCE.

| Model | CMU-MOSE | | | | | |
|---|---|---|---|---|---|---|
| | Acc-2 | Acc-5 | Acc-7 | F1 | Corr | MAE |
| EF-LSTM | 78.48 | 40.15 | 35.39 | 78.51 | 66.49 | 94.88 |
| LF-DNN | 78.63 | 38.05 | 34.52 | 78.63 | 65.84 | 95.48 |
| MFN | 78.87 | 40.47 | 35.83 | 78.9 | 67.02 | 92.68 |
| TFN | 79.08 | 39.39 | 34.46 | 79.11 | 67.33 | 94.73 |
| Mult | 79.71 | 42.68 | 36.91 | 79.63 | 70.22 | 87.99 |
| Ours | 83.44 | 41.23 | 35.42 | 83.21 | 72.38 | 88.42 |

TABLE IV
THE MULTIMODAL EXPERIMENTAL RESULTS ON THE CH-SIMS DATASET ARE COMPARED, WITH "OURS" REPRESENTING THE METHOD USED IN THIS PAPER. WE RAN EACH MODEL FIVE TIMES AND REPORTED THE AVERAGE RESULTS. FOR ALL METRICS EXCEPT MAE, HIGHER VALUES INDICATE BETTER PERFORMANCE.

| Model | CH-SIMS | | | | | |
|---|---|---|---|---|---|---|
| | Acc-2 | Acc-3 | Acc-5 | F1 | Corr | MAE |
| EF-LSTM | 69.37 | 54.27 | 21.23 | 56.82 | 5.45 | 59.07 |
| LF-DNN | 77.02 | 64.33 | 39.37 | 77.27 | 55.51 | 44.63 |
| MFN | 77.9 | 65.73 | 39.47 | 77.88 | 58.24 | 43.49 |
| TFN | 78.38 | 65.12 | 39.3 | 78.62 | 59.1 | 43.22 |
| Mult | 78.56 | 64.77 | 37.94 | 79.66 | 56.41 | 45.32 |
| Ours | 80.12 | 65.23 | 40.12 | 79.38 | 59.32 | 44.12 |

*3) Single-mode results :*

*a) Impact of modality:* In order to evaluate the impact of different modalities on sentiment analysis results, we conducted single-modal sentiment analysis comparison experiments on the CMU-MOSI and CH-SIMS datasets. The experimental results are shown in Fig. 5, where A, V, and T represent sentiment analysis of audio, visual, and text modalities, respectively.

Based on the results of single-modal experiments, significant performance differences are observed across different modalities on the CMU-MOSE dataset, with audio modalities performing the lowest and text modalities performing the highest. This is because text data typically contains more emotion-related information, such as emotion words and emoticons. Compared to audio and visual modalities, the performance improvement of the text modality is significant, approximately 10% for Acc-2 and about 15% for Acc-3. On the CH-SIMS dataset, there is little difference between the text and visual modalities, both showing excellent performance. This can be attributed to our preprocessing techniques for video, which effectively extract facial expression areas, thereby enhancing recognition effectiveness. Experimental results demonstrate that useful information related to sentiment analysis is unevenly distributed across different modalities, with the text modality making the most significant contribution to sentiment analysis.

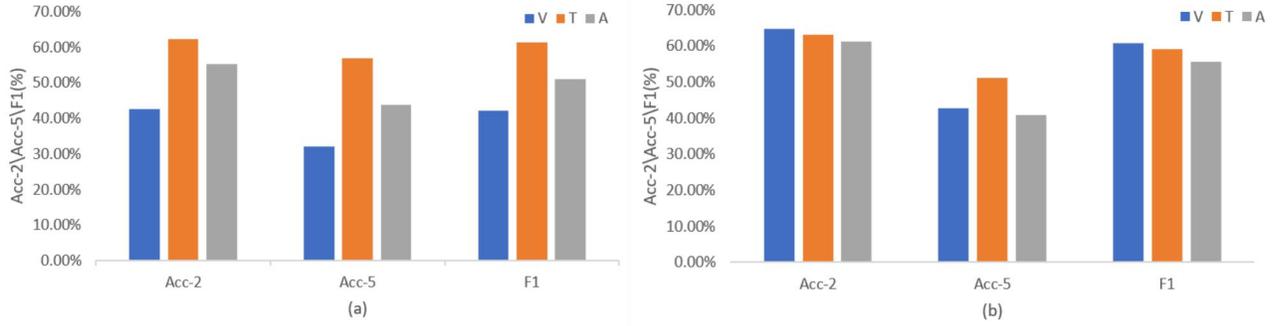

Fig. 5. Impact of modality. (a) CMU-MOSI. (b) CH-SIMS.

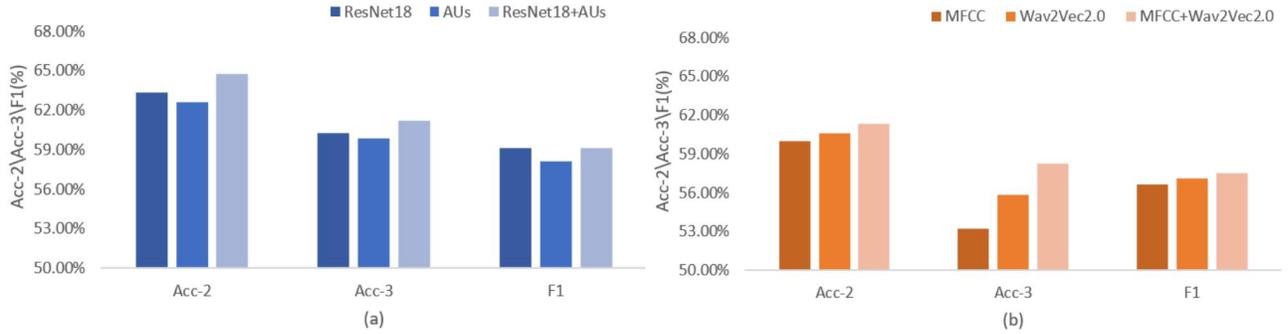

Fig. 6. Impact of features. (a) video. (b) audio.

*b) Impact of features:* To validate the performance impact of our proposed multi-channel features on single-modal emotion analysis, we conducted experiments on the visual and audio modalities separately using the CH-SIMS dataset. The experimental results are shown in Fig.6. In the figure, "ResNet18" denotes the backbone network for feature extraction, "AUs" represents extracted visual local features, and "+" indicates the fusion of two different channel features. For instance, "ResNet18+AUs" signifies the concatenation of these features in the spatial dimension.

It can be observed that according to the results of single-modal experiments, different features yield varying performance. On the CH-SIMS dataset, ResNet18 features outperform AUs features in the visual modality, showing an improvement of approximately 2% in Acc-2. Moreover, multi-channel features exhibit significantly better performance compared to single-channel features across all metrics. For the audio modality, while the overall experimental metrics show minimal differences, features based on MFCC+Wav2Vec2.0 outperform each individual single-channel feature (MFCC, Wav2Vec2.0) by approximately 1%-2% in Acc-2, Acc-3, and F1 scores. These results affirm the usefulness of the proposed multi-channel features, enhancing model performance and laying the groundwork for future modal feature fusion.

*4) Bimodal results :* This section will investigate bimodal emotion recognition models based on visual and audio, visual and text, and audio and text modalities. To further explore the impact of intra-modal features and inter-modal fusion features on the final recognition accuracy, this study compares experiments where inter-modal fusion features are directly used as inputs to the prediction module with experiments integrating intra-modal features and inter-modal fusion features as inputs to the prediction module. Experimental results are shown in Table V.

TABLE V
Bimodal Emotion Analysis Experimental Results

| Modalities | Features | Acc-3 | F1 |
|---|---|---|---|
| V+A | $F_{va}$ | 65.02 | 61.87 |
| | $F_v + F_a + F_{va}$ | 66.81 | 64.07 |
| V+T | $F_{vt}$ | 66.47 | 63.13 |
| | $F_v + F_t + F_{va}$ | 68.73 | 65.93 |
| A+T | $F_{at}$ | 64.83 | 62.03 |
| | $F_a + F_t + F_{at}$ | 66.19 | 63.50 |

For the bimodal emotion recognition models of visual and audio, visual and text, and audio and text, the experimental results on the CH-SIMS dataset from Table V. For the visual and audio emotion recognition model, feature column $F_{va}$ indicates using directly fused visual-audio features as inputs to the prediction module, achieving an accuracy of 65.02%. Feature column $F_v + F_a + F_{va}$ integrates visual features, audio features, and visual-audio fusion features as inputs, achieving a final accuracy of 66.81%, an improvement of 1.79% over the former. This validates the effectiveness of the proposed approach. For the visual and text emotion recognition model, feature column $F_{vt}$ denotes using directly fused visual-text features as inputs, achieving an accuracy of 66.47%. Feature column $F_v + F_t + F_{va}$ integrates visual features, text features, and visual-text fusion features as inputs, achieving a final accuracy of 68.73%, an improvement of 2.26% over the former. This confirms the effectiveness of the proposed approach. For the audio and text emotion recognition model, feature column $F_{at}$ represents using directly fused audio-text features as inputs, achieving an accuracy of 64.83%. Feature column $F_a + F_t + F_{at}$ integrates audio features, text features, and audio-text fusion features as inputs, achieving a final accuracy of 66.19%, an improvement of 1.36% over the former. This validates the effectiveness of the proposed approach.

Overall, in bimodal emotion analysis combining visual and audio, visual and text, and audio and text inputs, the model based on visual and text inputs achieves the highest recognition accuracy of 68.73%. This represents an improvement of 1.92% and 2.54% compared to models based on visual and audio, and audio and text inputs, respectively. This outcome is attributed to higher accuracy in single-modal emotion analysis models based on visual and text inputs compared to those based on audio inputs, confirming that superior single-modal feature extraction contributes to better performance in multimodal feature integration processes.

## V. Conclusion

In this paper, we focus on multimodal emotion recognition research using visual, audio, and text data. Experiments are conducted on the CMU-MOSE and CH-SIMS multimodal emotion datasets. The study is divided into two main parts: rich intra-modal feature extraction and effective inter-modal feature fusion. For the visual modality, the paper proposes global spatial feature extraction using the pre-trained ResNet18 network and local expression feature extraction based on AUs, enhancing the understanding and analysis of facial expressions. For the audio modality, traditional MFCC-based feature extraction and deep learning-based Wav2Vec2.0 feature extraction are employed, producing high-quality speech feature representations. The paper introduces a feature fusion module based on symmetric mutual promotion, integrating symmetric cross-modal attention and self-attention mechanisms to capture useful information from other modalities and model contextual information. By integrating intra-modal and inter-modal fusion features, the approach aims to retain intra-modal feature details while leveraging complementary inter-modal feature information. Experimental results validate the effectiveness and superiority of the proposed methodology.

## Acknowledgments

This work was supported by the Natural Science Foundation of China (62276242), National Aviation Science Foundation (2022Z071078001), CAAI-Huawei MindSpore Open Fund (CAAIXSJLJJ-2021-016B, CAAIXSJLJJ-2022-001A), Anhui Province Key Research and Development Program (202104a05020007), USTC-IAT Application Sci. & Tech. Achievement Cultivation Program (JL06521001Y), Sci. & Tech. Innovation Special Zone (20-163-14-LZ-001-004-01).

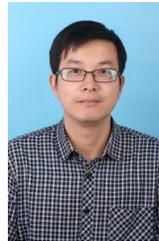

**Jun Yu** is currently an associate professor and laboratory director with the Department of Automation and the Institute of Advanced Technology, University of Science and Technology of China. His research interests are Multimedia Computing and Intelligent Robot. He has published 200+ journal articles and conference papers in TPAMI, IJCV, JMLR, TIP, TMM, TASLP, TCYB, TITS, TCSVT, TOMM, TCDS, ACL, CVPR, ICCV, NeurIPS, ICML, ICLR, MM, SIGGRAPH, VR, AAAI, IJCAI, etc. He has received 6 Best Paper Awards from premier conferences, including CVPR PBVS, ICCV MFR, ICME, FG, and won 60+ champions from Grand Challenges held in NeurIPS, CVPR, ICCV, MM, ECCV, IJCAI, AAAI. Email: harryjun@ustc.edu.cn.

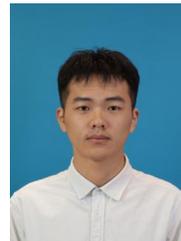

**Wangyuan Zhu** received the M.S. degree from the University of Science and Technology of China in 2024. His current research interests include multimodal sentiment analysis. Email: zhuwangyuan@mail.ustc.edu.cn.